\documentclass[10pt,twocolumn,letterpaper]{article}

\usepackage[pagenumbers]{cvpr} %

\definecolor{cvprblue}{rgb}{0.21,0.49,0.74}
\usepackage[pagebackref,breaklinks,colorlinks,allcolors=cvprblue]{hyperref}

\def\paperID{xxxx} %
\def\confName{CVPR}
\def\confYear{2025}

\title{Unsupervised Parameter Efficient Source-free Post-pretraining}

\author{Abhishek Jha$^1$\thanks{\url{email: abhishek.jha@esat.kuleuven.be}} 
 \hspace{0,5cm}  Tinne Tuytelaars$^1$ \hspace{0,5cm}  Yuki M. Asano$^2$\\
$^1$ESAT-PSI, KU Leuven \hspace{0,5cm} $^2$Fundamental AI Lab, University of Technology Nuremberg
}

\begin{document}
\maketitle
\begin{abstract}
  Following the success in NLP, the best vision models are now in the billion parameter ranges.
  Adapting these large models to a target distribution has become computationally and economically prohibitive. Addressing this challenge, we introduce UpStep, an Unsupervised Parameter-efficient Source-free post-pretraining approach, designed to efficiently adapt a base model from a source domain to a target domain: i) we design a self-supervised training scheme to adapt a pretrained model on an unlabeled target domain in a setting where source domain data is unavailable. Such source-free setting comes with the risk of catastrophic forgetting, hence, ii) we propose center vector regularization (CVR), a set of auxiliary operations that minimize catastrophic forgetting and additionally reduces the computational cost by skipping backpropagation in 50\% of the training iterations. Finally iii) we perform this adaptation process in a parameter-efficient way by adapting the pretrained model through low-rank adaptation methods, resulting in a fraction of parameters to optimize. We utilize various general backbone architectures, both supervised and unsupervised, trained on Imagenet as our base model and adapt them to a diverse set of eight target domains demonstrating the adaptability and generalizability of our proposed approach.
\end{abstract}
    
\section{Introduction}
\label{sec:intro}

Many recent advancements in computer vision like generative models \cite{ramesh2021zero, alayrac2022flamingo}, multi-modal models \cite{radford2021learning, li2022blip}, and learning generalized representations \cite{goyal2021self, dehghani2023scaling}, can be attributed to large models trained on large diverse datasets \cite{goyal2021self, thomee2016yfcc100m, sun2017revisiting}. These datasets typically contain natural images, collected by scraping the open-web, and hence models trained on such corpora tend to capture a generalized representation of the natural scenes. The motivation behind training these generalized representations is to transfer them to niche domains by adapting the parameters of these models for the respective target tasks. However, the size of recent visual models, reaching up to $22$B parameters \cite{dehghani2023scaling, riquelme2021scaling} makes them computationally and economically expensive to finetune, while being prohibitive to (re)train, for the larger part of the research community. Recent work in parameter-efficient training of large language models and large vision models like low-rank adaptation (LoRA) \cite{hu2021lora}, and prompt learning \cite{liu2023pre}, have shown effectiveness in adapting large language models by modifying a small subset of model parameters. While there has been some prior work on utilizing these adapters in domain adaptation \cite{zhu2024melo}, they are often limited to similar target domains.

Another challenge in adapting these models to a new target task is to generate enough samples from the target domain along with their respective task specific labels, as such labeling tasks can be expensive and prone to human error at scale \cite{northcutt2021pervasive, vasudevan2022does}. Self-supervision as the training task has shown to learn a generalized representation of the input distribution, that is easily transferred to a task in that domain. However, the use of such objectives for unsupervised target domain-adaptation is under-explored \cite{warmerdam2024self}.

Adapting a pretrained model to a new data distribution leads to forgetting of previous knowledge, also known as catastrophic forgetting \cite{de2021continual}. For large foundation models, the source domain is usually big. While including the entire source domain data is computationally prohibitive, including a subset requires coming up with sampling strategies, as in replay buffer methods \cite{de2021continual}. In this work, we consider the source-free setting, where the model doesn't have access to the source domain data.

We approach these challenges by proposing an Unsupervised Parameter-efficient Source-free post-preTraining, which we call UPST(ep), or UpStep. Specifically, in this paper:

i) we adapt a general model pretrained on a large source domain dataset, Imagenet \cite{krizhevsky2012imagenet} or WIT-400M \cite{radford2021learning} to a diverse set of target domains, using a self-supervised objective. We train a projector on top of the base model to project it into a lower dimensional space, where we perform online clustering.

ii) We propose a set of regularizers, collectively called Center vector regularization (CVR), based on variance maximization of the representation. We enforce this by minimizing the expected representation of the batch, called the center vector \cite{jha2024common}. CVR contains CV term as an auxiliary loss to the main self-supervised objective; a learning rate multiplier to promote the parameter updates when the variance is high and reduce its effect when the variance is low; and CV conditional gate, which skips the training when the variance decreases.

iii) We use low-rank adaptation technique to adapt our general source domain models to the target domain.

For evaluation, we employ a diverse set of datasets, showcasing the generalization capacity of our proposed approach, while building on different general pretrained models showcasing the adaptability of our model.

\begin{figure*}
    \centering
    \includegraphics[ width=0.9\linewidth]{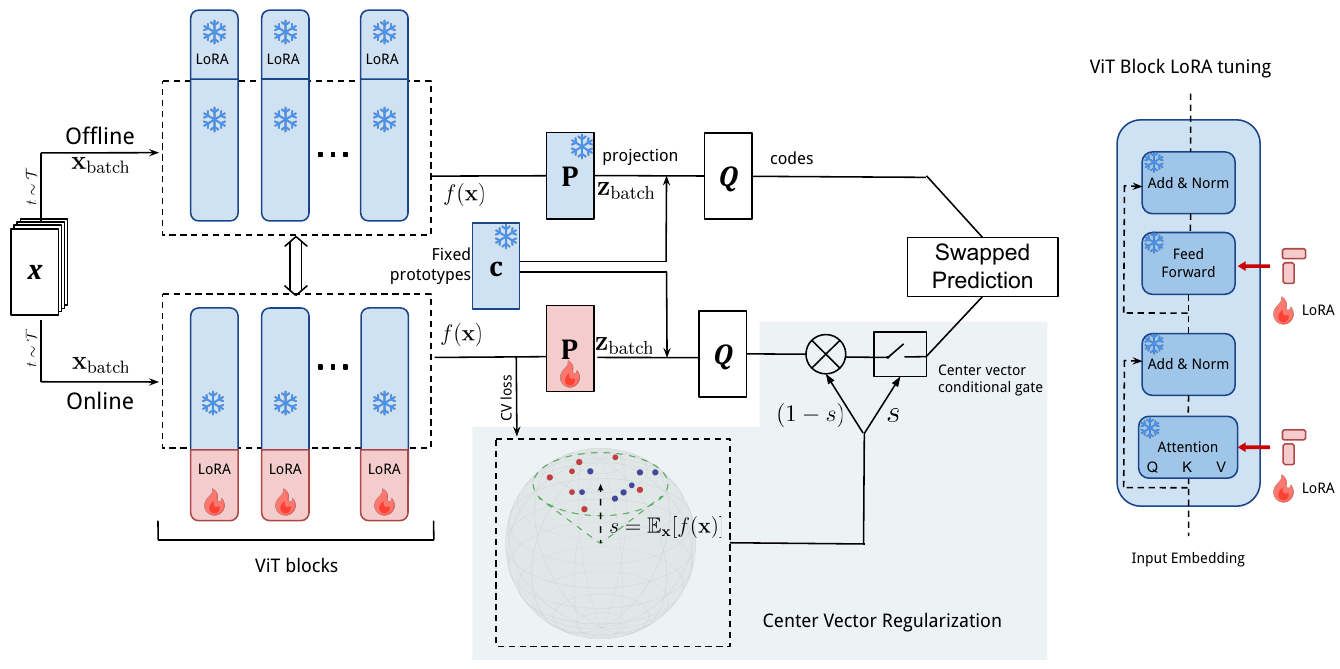}
    \caption{\textbf{Proposed UpStep architecture:} During training, we train the pass the augmented view of the input images through the online and offline streams. These online and offline are identical in architecture consisting of LoRA adapted pretrained Base models. The encoded representations are projected to the prototype space where an online clustering loss is applied. We apply an auxiliary loss, learning rate regularization and a gating mechanism to skip training for certain iterations conditioned upon the magnitude of center vector, as shown by the shaded region, Center vector regularization. For each layer in ViT model, we adapt the QKV matrices and the projection layers. During evaluation, we only use the LoRA adapted target domain base network in ensemble with the source domain base model.}
    \label{fig:UpStep_main}
\end{figure*}

\section{Method}
\label{sec:method}

Our proposed approach, UpStep, adapts a pretrained visual model to a target domain efficiently, without requiring access to source domain and without requiring target labels. 
The aim of our method is to extend the utility of generalized pretrained representations by tailoring them to a target distribution instead of a specific target task. Since the primary objective is domain adaptation rather than task-specific finetuning, we refer to this process as “post-pretraining,” in that the model's representation is the key output of our method, which can then be further refined for a given downstream task. The following sections provide a detailed description of our approach.

\subsection{Self-supervised Objective}
\label{sec:generalize}

We begin with a base model, pretrained on a general source domain, such as ImageNet \cite{krizhevsky2012imagenet}. Models pretrained with self-supervised learning (SSL) objectives have shown strong generalization capabilities; thus, we employ pretrained SSL encoders as our base model.

To further train this model on a target dataset, we utilize a two-stream (online-offline) self-supervised clustering-based approach, similar to SwAV \cite{caron2020unsupervised}. To maintain simplicity and efficiency, we employ fixed cluster centers, which are uniformly sampled on the unit hypersphere 
 \cite{caron2020unsupervised}, thereby reducing the training cost and minimizing the risk of model collapse. 

Let \( x \) represent a sample from the target domain. We generate two augmented views, \( x_s \) and \( x_t \). These views are processed by the online and offline streams respectively, which share identical architectures, each consisting of an encoder initialized with the base model, followed by a randomly initialized projector and a fixed set of cluster centers shared by both streams, as shown in \cref{fig:UpStep_main}

The encoded representations of \( x_s \) and \( x_t \), denoted as \( f(x_s) \) and \( f(x_t) \), are further projected to obtain \( z(x_s) \) and \( z(x_t) \), which are then assigned to clusters based on similarity to the fixed prototypes. The offline stream's cluster assignments \( \mathbf{p}_t \) are regularized using Sinkhorn-Knopp equipartitioning to enforce uniform distribution across clusters:

\begin{align}
\mathbf{p}_t = \text{SinkhornKnopp}(z(x_t) \cdot \mathbf{C})
\label{eq:sk}
\end{align}

where \( \mathbf{C} \) is the matrix of fixed cluster prototypes.

Finally, we compute a cross-entropy loss between the online and offline assignments:

\begin{align}
\mathcal{L}_{\text{CE}} = - \sum_k \mathbf{p}_t^{(k)} \log \mathbf{p}_s^{(k)},
\end{align}

where \( \mathbf{p}_s \) represents the online cluster assignments of \( z(x_s) \). This loss is backpropagated through the online stream to update the encoder and projector weights, while the offline stream is updated by the weights of the online stream after each iteration.

\subsection{Loss Regularization via Center vector}
\label{sec:loss_reg}

Further training a pretrained model on a target domain can result in unstable training loss, potentially degrading the pretrained knowledge. This is particularly an issue for small target distributions lacking sufficient diversity. To address this, we regularize the learning rate for the online network using batch statistics of the training samples.

Following \cite{zhang2022does, jha2024common}, we define the \textit{center vector (CV)}, \( \mathbf{s}_k \), as the average feature vector  after normalization across the batch,

\begin{align}
\mathbf{s}_k = \frac{1}{N} \sum_{t=1}^{N} \frac{f(x_s)}{\|f(x_s)\|},
\end{align}

where \( N \) is the batch size, and k indexes the training iteration. Training stability is inversely related to the magnitude of \( \|\mathbf{s}_k\| \), which measures the batch’s diversity, when the representations are lying on the unit hypersphere. Here, we define our auxiliary CV loss:

\begin{align}
    \mathcal{L}_{\text{CV}} = \big\|\|\mathbf{s}_k\|-\mathbf{s_\phi}\big\|
\end{align}
where $\mathbf{s_\phi}$ is a constant hyperparameter $\in [0,1]$, which controls the variance of the features on the unit sphere.

Furthermore, the learning rate \( \eta \) is adjusted for each batch iteration by CV learning rate regularization as follows:
\begin{align}
    \eta = \eta_0 \times (1 - \|\mathbf{s}_k\|),
\end{align}

where \( \eta_0 \) is the base learning rate. This learning rate regularization strategy, inspired by Catastrophic Forgetting Measurement (CFM) introduced in PADCLIP \cite{lai2023padclip}, helps stabilize training, especially for less diverse datasets. However, unlike CFM, which requires the access to source data, Center vector learning rate regularization only requires samples from target domain. 

Finally, we introduce a gating mechanism that allows gradients to backpropagate only if the center vector magnitude minimization condition is met. Specifically, the gradient is propagated when the current iteration’s center vector magnitude \( \|\mathbf{s}\| \) is lower than that of the previous iteration. Hence, our UpStep objective is given by:
\begin{align}
    \mathcal{L}_{\text{UpStep}} = 
    \begin{cases} 
      \big(\mathcal{L}_{\text{CE}} + \mathcal{L}_{\text{CV}}\big), & \text{if } \|\mathbf{s}_k\| < \|\mathbf{s}_{k-1}\| \\ 
      0, & \text{otherwise}
    \end{cases}
\end{align}
This encourages the network to update weights only when the variance of the samples in the feature space is high.

\subsection{Low-Rank Adaptation}
\label{sec:lora_adaptation}

To perform resource-efficient post-pretraining, we adapt the encoder using Low-Rank Adaptation (LoRA) layers applied to the Query-Key-Value (QKV) projection matrices in the self-attention and projection modules of each transformer layer.
This structured adaptation allows for flexible and efficient fine-tuning, as only essential components are modified, making it feasible to adapt large models like VFMs without updating the full set of parameters.

\subsection{Efficient Model Ensemble}
\label{sec:evaluation_method}
Finally, during evaluation, we use the ensemble of the base encoder model and the UpStep encoder model by concatenating the features from both models. This approach reduces the problem of catastrophic forgetting, which is especially critical when the target domain differs significantly from the source. Separately, we show that enforcing center-vector loss minimization during training helps reduce forgetting of the source domain knowledge, resulting in lesser dependence on the source domain model within the ensemble.

Let \( f_{\text{base}}(x) \) and \( f_{\text{UpStep}}(x) \) denote the feature representations from the base and UpStep models, respectively. The final feature vector used for evaluation is given by:

\begin{align}
    f_{\text{ensemble}}(x) = [f_{\text{base}}(x); f_{\text{UpStep}}(x)]
\end{align}

This concatenated representation leverages both models’ knowledge, balancing adaptation to the target domain and retaining essential information from the source domain.

\begin{table*}[ht]
\centering
\small
\setlength{\tabcolsep}{1.8pt}
\begin{tabular}{@{}l c cccccccccc@{}}
\toprule
\textbf{Method} & \textbf{\% Train Params} & \textbf{CIFAR-10} & \textbf{CIFAR-100} & \textbf{DTD} & \textbf{EuroSAT} & \textbf{Flowers102} & \textbf{Oxford Pets} & \textbf{SUN397} & \textbf{UCF101} & \textbf{Avg} \\
\midrule
\multicolumn{11}{l}{\textit{Pretrained model: $\text{ViT-B/32}_{\text{CLIP}}$}} \\
Base Model (k-NN) & 0\% & 0.909 & 0.694 & 0.666 & 0.858 & 0.818 & 0.768 & 0.687 & 0.753 & 0.769 \\
Base Model (lin) & 0\% & 0.946 & 0.795 & 0.723 & 0.950 & 0.907 & 0.887 & 0.755 & 0.828 & 0.849 \\
Mask (lin) \cite{warmerdam2024self} & 100\% & \textbf{0.971} & 0.834 & 0.738 & \textbf{0.978} & \textbf{0.973} & 0.891 & 0.668 & 0.815 & 0.858 \\
\midrule
UpStep (k-NN) & \textbf{4.5\%} & 0.960 & 0.801 & 0.714 & 0.965 & 0.885 & 0.826 & 0.712 & 0.790 & 0.832 \\
UpStep (lin) & \textbf{4.5\%} & 0.966 & \textbf{0.843} & \textbf{0.749} & 0.972 & 0.930 & \textbf{0.901} & \textbf{0.761} & \textbf{0.845} & \textbf{0.871} \\
\bottomrule
\end{tabular}
\caption{
\textbf{Comparison of Top-1 Accuracy and Training Parameters (Train Params) for Downstream Task.} 
We report the percentage of training parameters required for each method and the top-1 accuracy across eight image classification benchmarks, with k-NN classification and linear probing. The final column shows the average accuracy across all datasets.
}
\label{tab:results}
\end{table*}

\section{Experiments}
\subsection{Datasets}
\label{sec:datasets}

For the source domain, we use the ImageNet-1k dataset \cite{krizhevsky2012imagenet}, a widely recognized benchmark for pretraining visual models, consisting of 1.2 million training samples of natural images. For evaluating the target domain adaptation, we employ a diverse set of datasets. The CIFAR-10 and CIFAR-100 datasets \cite{krizhevsky2009learning} provide 50,000 training samples each, containing 10 and 100 classes respectively, and represent general object categories. EuroSAT \cite{helber2018introducing}, which is focused on satellite imagery, includes 27,000 training samples across 10 classes, while SUN397 \cite{xiao2010sun}, covering a broad range of scene types, has 19,850 training samples and 397 classes. Flowers102 \cite{nilsback2008automated}, representing various flower species, contains 1,020 training samples spanning 102 categories. UCF101 \cite{soomro2012ucf101}, focused on action recognition, has 13,000 samples with 101 classes. DTD \cite{cimpoi14describing}, designed for texture analysis, comprises 1,880 training samples across 47 classes, and Oxford Pets, which includes images of different pet breeds, has 200 training samples across 37 classes. This diverse selection of datasets allows us to comprehensively assess the generalization capability of our approach across different visual domains, testing its robustness and adaptability to new and varied distributions.

\subsection{Implementation Details}
\label{sec:implementation}

We initialize our model with ImageNet-pretrained models as the base. For the projector network, we employ a randomly initilized two-layer MLP with a hidden dimension of 2048 and an output dimension of 128, using ReLU as the activation function between layers. A total of 3000 cluster prototypes are utilized, which are uniformly sampled on the unit hypersphere to ensure balanced cluster distribution. The training is conducted with a batch size of 160, and unlike SwAV, we do not use a queue mechanism. For the center vector loss, we use $\mathbf{s}_\phi = 0.5$, and $\mathbf{s}_0$ is initilized as 1.

In applying the Sinkhorn-Knopp \cite{caron2020unsupervised} algorithm, we set \(\epsilon\) to 0.3 and perform 3 iterations to achieve equipartitioning of the clusters. Both the encoder and projector modules are trained with a learning rate of 0.03, optimized using the Adam optimizer for stable convergence.
For evaluation, we employ k-nearest neighbors (k-NN) classification, setting \( k = 20 \) to determine the neighborhood size.
We use the Hugging Face implementation of LoRA with the following parameters: the rank of the LoRA matrix is set to 16, while the scaling factor for the low-rank matrices is set to 1. Additionally, no dropout is applied to the LoRA matrix.

\subsection{Comparison against Baselines}
\label{sec:baselines_comparison}

To assess the quality of the target representation, we evaluate it on the downstream task of classification. Since our approach builds upon the base model, the primary baseline is the ImageNet-1k pretrained model. Additionally, we compare our method against the approach proposed by Warmerdam \etal \cite{warmerdam2024self}, which trains a unsupervised masking network on top of the base model to adapt it to the unlabeled target domain, providing a strong baseline for the comparison.

As shown in \cref{tab:results}, our method achieves competitive performance across all datasets compared to both the baselines, while offering the additional advantages of much reduced training parameters when compared against Warmerdam \etal \cite{warmerdam2024self}.

\subsection{Evaluating Different Base Models}
\label{sec:different_base_models}

We evaluate our method on a range of base models to assess its generalization capability across diverse pretrained representations, while using k-NN classification evaluation. Specifically, we use Imagenet-1k Supservised ViT-Base\/16 \cite{dosovitskiy2020image}, DINO-pretrained models \cite{caron2021emerging}: ViT-Small\/16 and ViT-Base\/16, CLIP ViT-Base\/32 \cite{radford2021learning}, and DINOv2R \cite{darcet2023vitneedreg} as our base models. As shown in \cref{tab:different_bases}, our method demonstrates strong performance across different datasets and pretrained models, often outperforming the baselines. These results demonstrate that UpStep can be effectively applied to different based models pretrained with supervised, self-supervised or multimodal objectives.

\begin{table*}[ht]
\centering

\small
\setlength{\tabcolsep}{1.8pt}
\begin{tabular}{@{}l c c cccccccc@{}}
\toprule
\textbf{Method} & \textbf{\% Training Parameters} & \textbf{CIFAR-10} & \textbf{CIFAR-100} & \textbf{DTD} & \textbf{EuroSAT} & \textbf{Flowers102} & \textbf{Oxford Pets} & \textbf{SUN397} & \textbf{UCF101} \\
\midrule

\multicolumn{10}{l}{\textit{Pretrained model: $\text{ViT-Base}_{\text{Supervised}}$}} \\
Base Model & 0\% & 93.77 & 77.26 & 66.54 & 91.30 & 95.10 & 90.59 & 64.10 & 73.96 \\
UpStep (our) & 5.1\% & \textbf{97.81} & \textbf{89.56} & \textbf{72.60} & \textbf{97.0} & \textbf{98.66} & \textbf{92.12} & \textbf{67.64} & \textbf{79.56} \\
\midrule

\multicolumn{10}{l}{\textit{Pretrained model: DINO ViT-Small}} \\
Base Model & 0\% & 95.03 & 79.72 & \textbf{71.27} & 95.39 & 80.19 & 90.81 & 59.26 & 73.16 \\
UpStep (our) & 7.5\% & \textbf{96.96} & \textbf{85.11} & 70.26 & \textbf{97.80} & \textbf{89.26} & \textbf{91.52} & \textbf{59.74} & \textbf{75.23} \\
\midrule

\multicolumn{10}{l}{\textit{Pretrained model: DINO ViT-Base}} \\
Base Model & 0\% & 96.43 & 82.87 & \textbf{71.12} & 95.19 & 82.69 & 89.64 & 61.09 & 75.41 \\
UpStep (our) & 5.1\% & \textbf{97.97} & \textbf{87.51} & 71.01 & \textbf{97.61} & \textbf{89.59} & \textbf{90.40} & \textbf{61.55} & \textbf{77.05} \\
\midrule

\multicolumn{10}{l}{\textit{Pretrained model: DINOv2R}} \\
Base Model & 0\% & 97.39 & 86.09 & 75.05 & 92.0 & 99.41 & 93.24 & 70.2 & 79.3 \\
UpStep (our) & 10.6\% & \textbf{98.36} & \textbf{89.94} & \textbf{76.64} & \textbf{97.06} & \textbf{99.51} & \textbf{93.45} & \textbf{71.59} & \textbf{81.25} \\
\bottomrule
\end{tabular}
\caption{
\textbf{Performance Comparison of Base Model and UpStep on Various Pretrained Models.} 
We report the top-1 accuracy for eight image classification benchmarks using different pretrained base models: Supervised ViT-Base/16 \cite{dosovitskiy2020image}, DINO ViT-Small/16, DINO ViT-Base/16 \cite{caron2021emerging}, and DINOv2R \cite{darcet2023vitneedreg}. UpStep (our) demonstrates improved performance across all datasets.
}
\label{tab:different_bases}
\vspace{-1.3em}
\end{table*}

\subsection{Ablation Study}
\label{sec:ablation}

We perform an ablation study to understand the importance of different components in our method. Specifically, we ablate the center vector regularization (CV Loss, learning rate regularization, and CV conditioned gating), and the ensemble of the base model and the UpStep model. As shown in \cref{tab:component_ablation}, each design choice contributes significantly to the performance of our method. 
It should be noted that while adding CV conditioned gating reduces the performance of Base only naive adaptation from $85.02\%$ to $84.9\%$, it reduces the total number of training iterations by 50\%, see \cref{{sec:reduction_in_training_time}}, resulting in a more efficient training scheme. Most performance gain can be attributed to CV Loss and ensembel, both of these operations minimize the catastrophic-forgetting, as discussed in detail in \cref{sec:center_vector_catastrophic_forgetting} and \cref{sec:evaluation_method} respectively.
Learning rate regularization also promote minimization of center vector magnitude, hence results in an improvement over naive Base only model.
Overall, we observe that the center vector regularization and ensemble is crucial for faster convergence and minimization of catastrophic forgetting of the source domain knowledge.

\begin{table}[h]
\centering

\small
\setlength{\tabcolsep}{3pt} %
\begin{tabular}{@{}lcccccc@{}}
\toprule
\textbf{Config} & \textbf{Base} & \multicolumn{3}{c}{\textbf{CV}} & \textbf{Ensemb} & \textbf{Acc \%} \\
\cmidrule(lr){3-5}
 & & \textbf{LR Reg} & \textbf{Cond} & \textbf{Loss} & & \\
\midrule
Base Only & \checkmark &  &  &  &  & 85.02 \\
Base + CV Cond & \checkmark &  & \checkmark &  &  & 84.90 \\
Base + CV LR Reg & \checkmark & \checkmark &  &  &  & 85.28 \\
Base + CV Loss & \checkmark &  &  & \checkmark &  & 87.51 \\
Base + CV (All) & \checkmark & \checkmark & \checkmark & \checkmark &  & 87.75 \\
UpStep & \checkmark & \checkmark & \checkmark & \checkmark & \checkmark & \textbf{89.26} \\
\bottomrule
\end{tabular}
\caption{
\textbf{Ablation Study of UpStep Components.} 
We evaluate the impact of each component, including the LoRA base model (Base), CV learning rate regularization (LR Reg), CV conditioned  gating (Cond), CV loss (Loss), and ensemble (Ensemb), on Flowers102 dataset. Checkmarks indicate the components present in each configuration, with accuracy (Acc) reported for each combination.}
\label{tab:component_ablation}
\vspace{-2.0em}
\end{table}

\subsection{Can we make the number of trainable parameters even lower?}
\label{sec:low_rank}

Our method uses Low-Rank Adaptation (LoRA) to efficiently adapt weight matrices by training low-rank matrices \( A \) and \( B \) with an intermediate rank, reducing trainable parameters compared to full fine-tuning. Vector-based Random Matrix Adaptation (VeRA) \cite{kopiczkovera} builds on LoRA by freezing a single pair of low-rank matrices, shared across all layers, and introducing layer-specific trainable scaling vectors. These scaling vectors enable each layer to be adjusted independently while sharing the same low-rank basis, drastically lowering trainable parameters. 

Like LoRA, VeRA’s trained scaling vectors can be merged into the original weights after training, adding no inference latency. When employed our method as the adapter, VeRA can further reduce parameter requirements from $7.9\%$ to $3.1\%$ for ViT-Base/16 model), thereby enhancing efficiency for large models or resource-limited settings with comparative performance. We compare the performances of UpStep-LoRA and UpStep-Vera versions of our approach in \cref{tab:upstep_lora_vera}.

\begin{table}[ht]
\centering
\small
\setlength{\tabcolsep}{4pt}
\begin{tabular}{@{}lcc@{}}
\toprule
\textbf{Dataset} & \textbf{UpStep-LoRA} & \textbf{UpStep-VeRA} \\
\midrule
\textbf{Training Parameters} & 7.9\% & 3.1\% \\
\textbf{Parameters to store} & 5.5\% & 0.03\% \\
\midrule
CIFAR-10        & 96.96 & 96.51 \\
CIFAR-100       & 85.11 & 83.07 \\
DTD             & 70.26 & 70.10 \\
EuroSAT         & 97.80 & 97.07 \\
Flowers102      & 89.26 & 88.06 \\
Oxford Pets     & 91.52 & 91.82 \\
SUN397          & 59.74 & 59.27 \\
UCF101          & 75.23 & 73.98 \\
Avg             & 83.24 & 82.49 \\
\bottomrule
\end{tabular}
\caption{
\textbf{Performance Comparison of UpStep-LoRA and UpStep-VeRA Across Datasets.} 
We evaluate the accuracy of UpStep with LoRA and VeRA on various datasets, highlighting the effectiveness of each approach in adapting to different domains with efficient parameter usage. Avg shows average performance over all datasets.
}
\label{tab:upstep_lora_vera}
\vspace{-1.3em}
\end{table}

\section{Analysis and Discussion}

\begin{figure*}[t]
\centering
\includegraphics[width=0.9\linewidth]{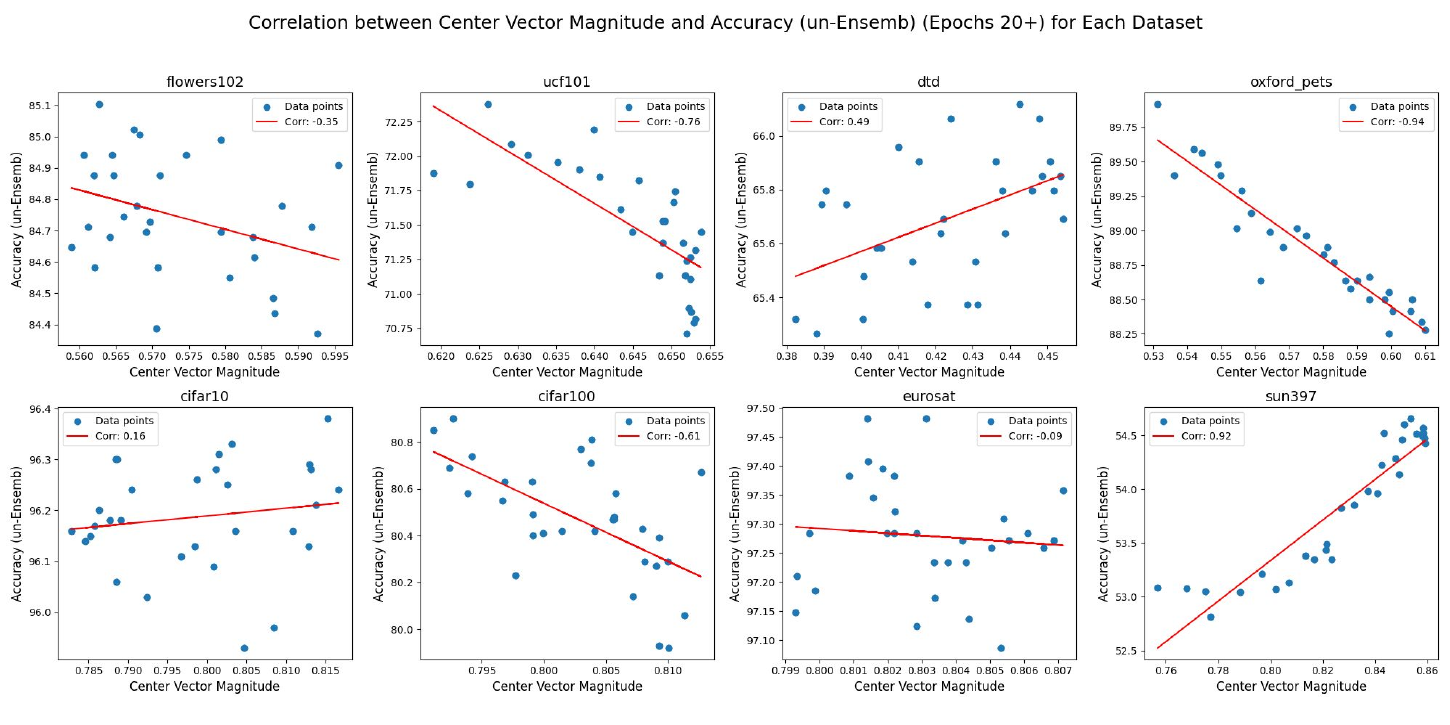}
\caption{
\textbf{Impact of Center Vector Magnitude on Model Performance.} 
Higher center vector magnitudes correlate with reduced k-NN accuracy for the majority of the datasets, underscoring the stabilizing role of center vector regularization.
}
\label{fig:correlation_cv_and_performance}
\end{figure*}

\begin{figure}[h]
\centering
\includegraphics[width=\linewidth]{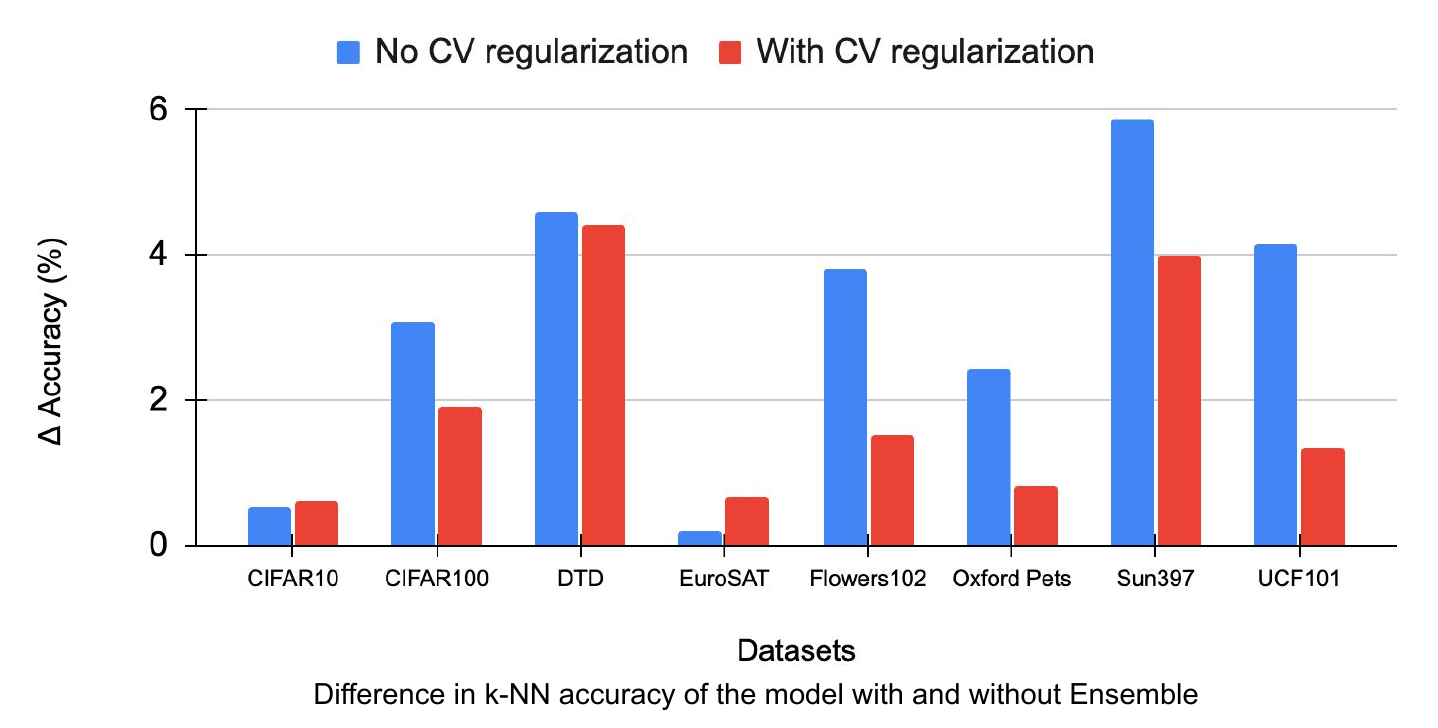}
\caption{
\textbf{Impact of Center Vector Regularization on Catastrophic Forgetting.}
Bar plots shows the difference in accuracy between the ensembled model, and un-ensembled model. Red bars are corresponding to UpStep model with center vector (CV) regularization, while the Blue bars represent the ablated version of UpStep without the CV regularization.
}
\label{fig:cv_and_forgetting}
\end{figure}

\subsection{Correlation between performance and Center vector magnitude}
\label{sec:correlation_performance_center_vector}

One of the key components of our method is the center vector regularization. Hence it is important to understand  how the center vector behaves during post-pretraining on the target domain. To this end we analyze the training time center vector magnitude and the k-NN classification performance of the model without the center vector loss and ensemble on the target domain test set. We observe that, in the absence of center vector loss, the center vector magnitude increases during training which is in line with similar findings on the self-supervised pretaining task by Jha \etal \cite{jha2024common}. They attribute stability of the self-supervised pretraining to center vector magnitude minimization, which explains our observation for the under-performance of the model on the target domain in the absence of center vector loss.
To further validate this, we also analyze the correlation between the center vector magnitude and the k-NN classification performance of the model on the target domain test set. From \cref{fig:correlation_cv_and_performance} we observe a negative correlation between center vector magnitude and k-NN classification performance in five out of eight datasets, suggesting that lower center vector magnitude often aligns with improved training stability during post-pretraining. However, for the remaining three datasets, a positive correlation is observed, indicating that the relationship may vary based on specific target domain characteristics.

\begin{figure}[h]
\centering
\includegraphics[width=\linewidth]{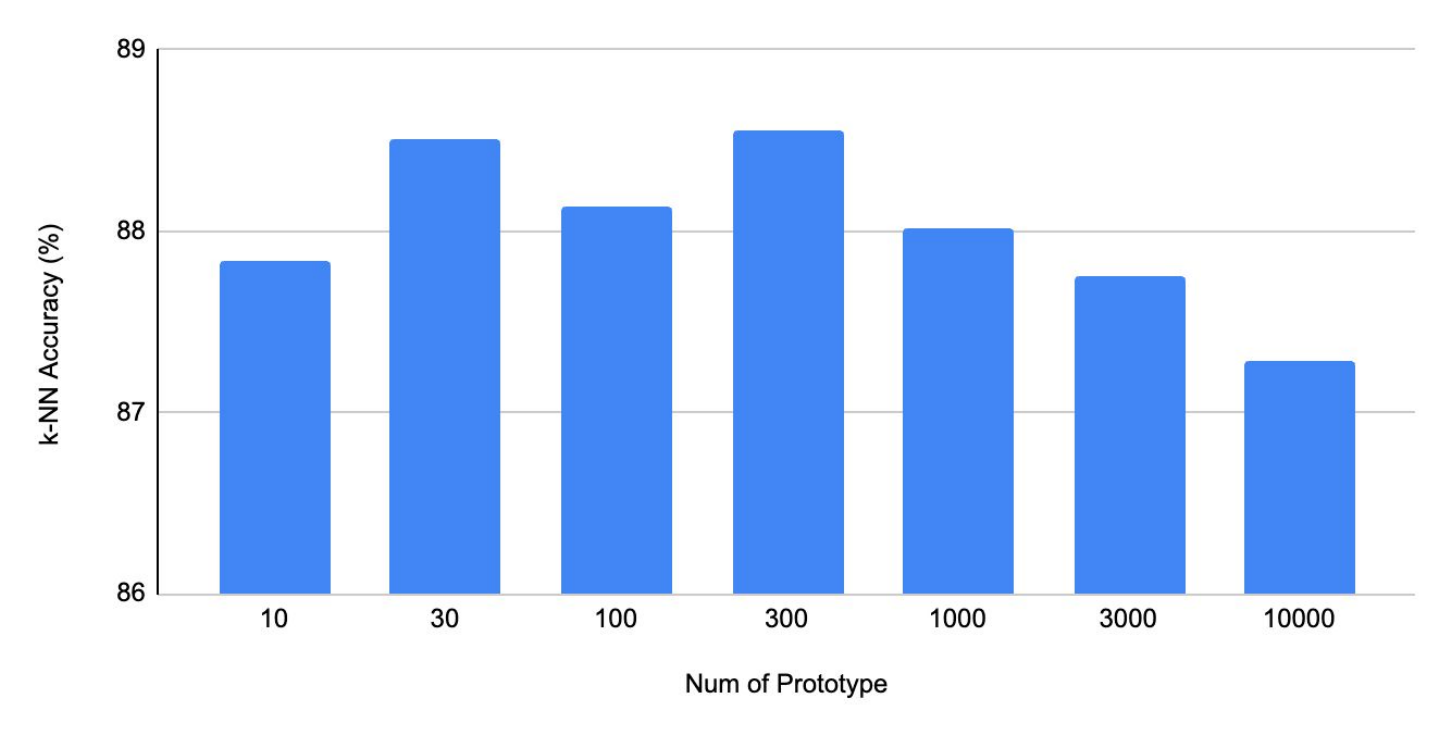}
\caption{
\textbf{Effect of Number of Prototypes on Model Performance.} 
Performance of the model with varying numbers of prototypes in the non-ensemble setting, on Flowers102 dataset \cite{nilsback2008automated}.
}
\label{fig:num_prototypes_effect}
\end{figure}

\begin{figure*}[th]
\centering
\includegraphics[width=\linewidth]{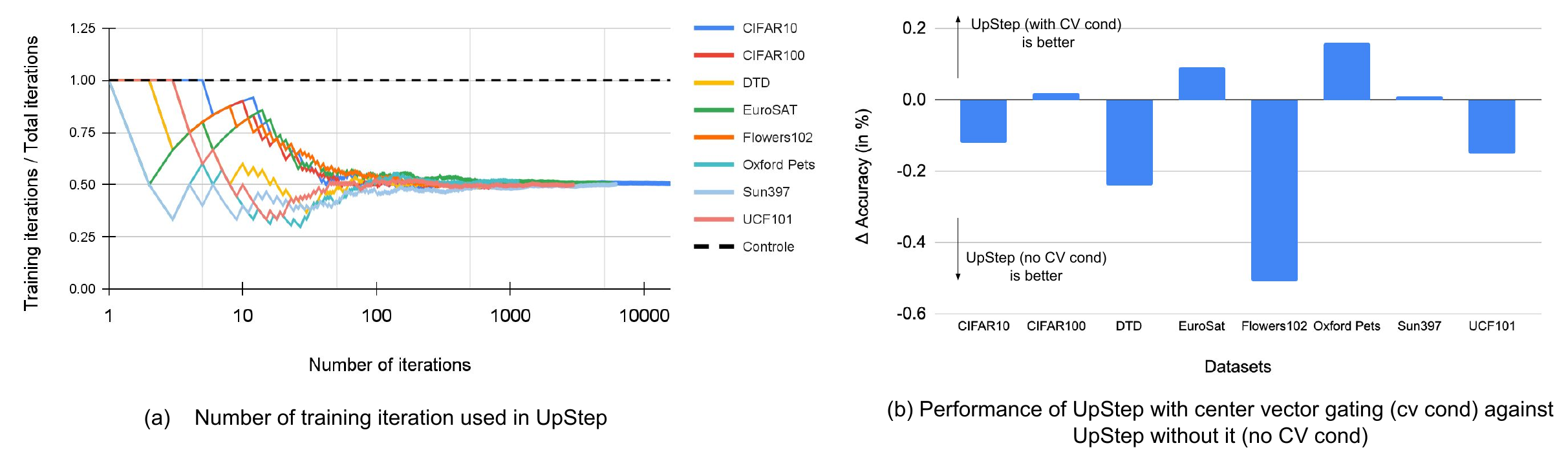}
\caption{
\textbf{Training Time Reduction with Center Vector Conditional Training.} 
(a) Average reduction in training time across datasets. (b) Performance comparison between UpStep with and without center vector (CV cond) conditioned gating. With a comparable performance over the dataset, Upstep with CV conditioned gating reduces the number of training iterations (backpropagation) to 50\%.
}
\label{fig:training_time_reduction}
\end{figure*}

\subsection{Center Vector and Catastrophic Forgetting}
\label{sec:center_vector_catastrophic_forgetting}

To minimize the effect of the catastrophic forgetting of source domain knowledge, we employ an ensemble of the post-pretrained model with the original source-domain model. By retaining the source model parameters, we protect against the negative impact of adapting to smaller or less diverse target domains. Interestingly, we observe that minimizing center vector magnitude also reduces catastrophic forgetting, thereby decreasing the ensemble’s reliance on the source domain model.

To examine this, we compare the classification performance of models trained with and without center vector regularization and analyze their performance in both ensemble and non-ensemble setups. Our results, \cref{fig:cv_and_forgetting} indicate that, without center vector regularization, the performance gap between the ensembled and non-ensembled models is larger, highlighting the role of center vector regularization in preserving source domain knowledge. For baseline performances that are saturated, i.e. ($~97\%$), in the case of CIFAR 10 and EuroSAT, we do not observe this phenomenon.

\subsection{Effect of the Number of Prototypes}
\label{sec:number_of_prototypes}

In our experiments, we use $3000$ prototypes following the design choice in SwAV \cite{caron2020unsupervised}. Here, we analyze the effect of varying the number of prototypes on the UpStep model's performance in the non-ensemble setting, as this model participates in the loss minimization and is directly affected by the prototype design choice. As shown in \cref{fig:num_prototypes_effect}, we observe that a very low number of prototypes results in a lower performance, likely because fewer prototypes cannot capture the diversity of the target domain, causing samples of different concepts to compete for the same prototypes. Conversely, an excessive number of prototypes creates a high-dimensional space that can easily overfit the target domain. Optimal performance is achieved with a moderate number of prototypes, balancing diversity representation with overfitting prevention.

\subsection{Reduction in Training Time}
\label{sec:reduction_in_training_time}

Our method incorporates center vector conditional batch training, where training occurs only if the current batch’s center vector is lower than that of the previous batch. This strategy reduces training time by skipping non-informative updates. On average, training time is reduced by 50\% across datasets, as shown in \cref{fig:training_time_reduction} (a), with comparative performance, as shown in \cref{fig:training_time_reduction} (b). Combined with LoRA, this approach not only reduces trainable parameters but also improves computational efficiency for post-pretraining.

\section{Related Work}
\label{sec:related_work}

The rapid expansion of vision model scales has accelerated research into parameter-efficient adaptation methods, aimed at overcoming the computational and economic barriers in adapting large models to novel domains. This section explores significant advancements in unsupervised and source-free domain adaptation, frozen network adaptation, foundational model adaptation, adversarial learning for feature alignment, self-supervised learning, and masking techniques. These methodologies contribute to the development of our UpStep approach.
\smallskip

\textbf{Unsupervised and Source-Free Domain Adaptation.} Unsupervised domain adaptation (UDA) methods seek to transfer knowledge from a labeled source domain to an unlabeled target domain by reducing domain shift through techniques like instance re-weighting and feature adaptation \cite{jiang2007instance, wang2017instance, yao2010boosting, gopalan2011domain, gong2012geodesic, sun2015subspace}. Source-free domain adaptation (SFDA) extends UDA by addressing scenarios where source data is inaccessible, e.g.~due to privacy constraints. SFDA methods adapt a pretrained source model to the target domain without access to source samples, utilizing strategies like self-supervised pseudo-labeling and knowledge extraction from the source model \cite{liang2020we, qiu2021source, kim2020progressive}. Our approach follows this source-free paradigm, adapting a pretrained model to a target domain without requiring labeled target data, thus addressing source-data availability constraints. Moreover, unlike SFDA approaches, we consider larger domain difference through a diverse set of target datasets.
\smallskip

\textbf{Frozen Network Adaptation.} Recent advancements in parameter-efficient adaptation techniques, that preserve core network weights, have shown promising results for domain adaptation without excessive computational costs. Low-rank adaptation (LoRA), for example, integrates trainable low-rank layers into pretrained models, enabling efficient adaptation with minimal parameter modification \cite{hu2021lora}. Other notable methods, including feature adapters \cite{gao2024clip, zhang2022tip}, bias tuning \cite{cai2020tinytl, zaken2021bitfit}, and visual prompting \cite{bahng2022exploring, jia2022visual}, achieve domain specialization by modifying only selected layers. Our work builds on these advancements by employing LoRA for parameter-efficient adaptation of the pretrained models to the target domain.

\smallskip
\textbf{Vision Foundation Models (VFMs).} Vision Foundation Models, such as DinoV2 \cite{oquab2023dinov2} and MAE \cite{he2021masked}, exhibit exceptional generalization across diverse tasks, largely due to their self-supervised pretraining on large-scale datasets like ImageNet \cite{oquab2023dinov2, he2021mae, grill2020bootstrap, chen2020improved}. Such models, pretrained on extensive data distributions, serve as robust starting points for domain-specific adaptation \cite{cong2022satmae, reed2022scale, moutakanni2024advancing, zhang2023text2seg}. Our method leverages VFMs pretrained on ImageNet~\cite{krizhevsky2012imagenet}, LVD-142M \cite{oquab2023dinov2}, and  WIT-400M \cite{radford2021learning}, while adapting it to target domains through a LoRA-enhanced self-supervised approach to optimize the domain-specific performance.

\smallskip

\textbf{Self-supervised Learning on Restricted Domains.} Self-supervised learning (SSL) techniques have shown notable success in extracting robust representations from unlabeled data \cite{caron2021emerging, caron2020unsupervised, he2019momentum, fang2021seed}. In our UpStep approach, SSL plays a critical role in capturing the semantics of the target domain ensuring label-free domain adaptation.

SSL-based knowledge distillation has demonstrated enhanced model performance on downstream tasks in vision and natural language processing by continuing training on unlabeled target datasets \cite{reed2022self, gururangan2020don, howard2018universal}. However, for these models the source and target distributions typically lie in similar domains, which is not a constraint for our approach. 
Moreover, these models train all the parameters of the pretrained model on the target dataset, making them resource-expensive, while our approach employs parameter efficient adaptation.
\smallskip

\textbf{Sparsity and Low-Rank.} Masking and sparsity methods enable efficient domain adaptation by selectively activating key subnetworks or by pruning unnecessary parameters. For instance, the pass-through trick activates specific subnetworks within pretrained models, enhancing adaptation efficiency without complete retraining \cite{mallya2018piggyback, mallya2017packnet, ramanujan2020s, wortsman2020supermasks}. Pruning techniques for convolutional neural networks (CNNs) and vision transformers further support adaptation with minimal overhead \cite{sanh2020movement, lagunas2021block}. Although UpStep does not directly incorporate masking, these techniques highlight the advantages of creating a sparse, efficient low-rank adaptation strategy.

In conclusion, we present a parameter-efficient solution for domain adaptation by integrating LoRA with self-supervised objective, offering a novel, label-free and cost-effective approach for adapting large vision models to domain-specific data.

\section{Conclusion}
\label{sec:conclusion}

In this work, we introduced UpStep, an efficient, unsupervised, source-free post-pretraining approach for adapting large pretrained vision models to new target domains without requiring source domain data or target labels. Our approach leverages three main contributions: first, a self-supervised training scheme to enable effective domain adaptation; second, center vector regularization (CVR) to mitigate catastrophic forgetting and to reduce training time by skipping backprpagation for 50\% of the training iterations; and third, in the unsupervised context, a parameter-efficient fine-tuning strategy through Low-Rank Adaptation (LoRA) that significantly minimizes the number of trainable parameters.

Extensive evaluations across diverse datasets and pretrained models demonstrate that UpStep achieves competitive or superior performance, often outperforming baselines while maintaining efficiency. We observed that center vector regularization helps retain discriminability of features, as shown in the analysis of correlation between center vector magnitude and model performance. Additionally, the results indicate that UpStep's flexibility allows it to be applied effectively across various architectures and target domains, further establishing its adaptability. While our method generally performs well, we observed that the correlation between center vector magnitude and classification performance varied across certain datasets, indicating an area for further investigation into the relationship between feature variance and domain-specific adaptation needs.

{
    \small
    \bibliographystyle{ieeenat_fullname}
    \bibliography{main}
}

\end{document}